\documentclass[12pt]{article}

\usepackage{amsmath}
\usepackage{times}
\usepackage{graphicx}
\usepackage{color}
\usepackage{multirow}
\usepackage[round]{natbib}
\usepackage{rotating}
\usepackage{bbm}
\usepackage{latexsym}
\usepackage{main}

\usepackage[hidelinks]{hyperref}
\usepackage{url}
\usepackage{times}
\usepackage{amssymb}
\usepackage{amsmath}
\usepackage{amsthm}
\usepackage{graphicx}
\usepackage{multicol}
\usepackage{booktabs}
\usepackage{resizegather}
\usepackage{caption}

\newcommand\numberthis{\addtocounter{equation}{1}\tag{\theequation}}

\newcommand{\captionfonts}{\normalsize}

\makeatletter  
\long\def\@makecaption#1#2{%
  \vskip\abovecaptionskip
  \sbox\@tempboxa{{\captionfonts #1: #2}}%
  \ifdim \wd\@tempboxa >\hsize
    {\captionfonts #1: #2\par}
  \else
    \hbox to\hsize{\hfil\box\@tempboxa\hfil}%
  \fi
  \vskip\belowcaptionskip}
\makeatother   

\begin{document}
\hspace{13.9cm}1

\ \vspace{20mm}\\

\noindent{\LARGE Dynamic Neural Turing Machine with\\ Continuous and Discrete Addressing Schemes}

\ \\
{\bf \large Caglar Gulcehre$^{\displaystyle 1}$, Sarath Chandar$^{\displaystyle 1}$, Kyunghyun Cho$^{\displaystyle 2}$, Yoshua Bengio$^{\displaystyle 1}$}\\
{$^{\displaystyle 1}$University of Montreal, {\small \tt name.lastname@umontreal.ca}}\\
{$^{\displaystyle 2}$New York University, {\small \tt name.lastname@nyu.edu}}\\
%


\thispagestyle{empty}
\markboth{}{NC instructions}
\ \vspace{-0mm}\\
%
\begin{center} {\bf Abstract} \end{center}
We extend neural Turing machine (NTM) model into a {\it
    dynamic neural Turing machine} (D-NTM) by introducing a trainable memory
    addressing scheme. This addressing scheme maintains for each memory cell two separate
    vectors, {\it content} and {\it address} vectors. This allows the D-NTM to
    learn a wide variety of location-based addressing strategies including both
    linear and nonlinear ones. We implement the D-NTM with both continuous,
    differentiable and discrete, non-differentiable read/write mechanisms. We investigate
    the mechanisms and effects of learning to read and  write into a memory through experiments 
    on Facebook bAbI tasks using both a {\bf feedforward} and {\bf GRU}-controller. The D-NTM
    is evaluated on a set of Facebook bAbI tasks and shown to outperform NTM and LSTM baselines.
    We have done extensive analysis of our model and different variations of NTM on bAbI task.
    We also provide further experimental results on sequential $p$MNIST, Stanford Natural 
    Language Inference, associative recall and copy tasks.

\section{Introduction}

Designing of general-purpose learning algorithms is one of the long-standing goals
of artificial intelligence. Despite the success of deep learning in this area
(see, e.g., \citep{Goodfellow-et-al-2016-Book}) there are still a set of complex tasks
that are not well addressed by conventional neural network based models. Those tasks often
require a neural network to be equipped with an explicit, external memory in
which a larger, potentially unbounded, set of facts need to be stored. They
include, but are not limited to, episodic question-answering~\citep{weston2014memory,hermann2015teaching,hill2015goldilocks}, compact algorithms~\citep{zaremba2015learning}, dialogue \citep{serban2016building,vinyals2015neural} and video caption generation~\citep{yao2015capgenvid}.

Recently two promising approaches that are based on neural networks for this type of tasks
have been proposed. Memory networks~\citep{weston2014memory} explicitly store all
the facts, or information, available for each episode in an external memory (as
continuous vectors) and use the attention-based mechanism to index them when
returning an output. On the other hand, neural Turing
machines (NTM,~\citep{graves2014neural}) read each fact in an episode and decides
whether to read, write the fact or do both to the external, differentiable memory.

A crucial difference between these two models is that the memory network does
not have a mechanism to modify the content of the external memory, while the NTM
does. In practice, this leads to easier learning in the memory network, which in
turn resulted in that it being used more in realistic tasks~\citep{bordes2015large,dodge2015}.
On the contrary, the NTM has mainly been tested on a series of small-scale,
carefully-crafted tasks such as copy and associative recall. However, NTM is more expressive, precisely 
because it can store and modify the internal state of the network as 
it processes an episode and we were able to use it without any modifications on the model 
for different tasks.

The original NTM supports two modes of addressing (which can be used
simultaneously.) They are content-based and location-based addressing. We notice
that the location-based strategy is based on linear addressing. The distance
between each pair of consecutive memory cells is fixed to a constant. We address
this limitation, in this paper, by introducing a learnable address vector for
each memory cell of the NTM with least recently used memory addressing mechanism, 
and we call this variant a {\it dynamic neural Turing machine} (D-NTM).

We evaluate the proposed D-NTM on the full set of Facebook bAbI
task~\citep{weston2014memory} using either {\bf continuous}, differentiable attention or
{\bf discrete}, non-differentiable attention~\citep{rlntm} as an addressing strategy. Our
experiments reveal that it is possible to use the discrete, non-differentiable
attention mechanism, and in fact, the D-NTM with the discrete attention and GRU controller
outperforms the one with the continuous attention. 
We also provide results on sequential $p$MNIST, Stanford Natural Language Inference (SNLI)
task and algorithmic tasks proposed by \citep{graves2014neural} in order to 
investigate the ability of our model when dealing with long-term dependencies.

We summarize our contributions in this paper as below,

\begin{itemize}
    \item We propose a variation of neural Turing machine called a dynamic
        neural Turing machine (D-NTM) which employs a learnable and location-based
        addressing.
    \item We demonstrate the application of neural Turing machines on more natural
        and less toyish tasks, episodic question-answering, natural language entailment, digit classification 
        from the pixes besides the toy tasks. We provide a detailed analysis of our model on the bAbI task.
    \item We propose to use the discrete attention mechanism and empirically show
        that, it can outperform the continuous attention based addressing for episodic QA task.
    \item We propose a curriculum strategy for our model with the feedforward controller
    and discrete attention that improves our results significantly.
\end{itemize}

In this paper, we avoid doing architecture engineering for each task we work on and focus on pure model's overall performance on each without task-specific modifications on the model. In that respect, we mainly compare our model against similar models such as NTM and LSTM  without task-specific modifications. This helps us to better understand the model's failures.

The remainder of this article is organized as follows. In Section 2, we describe the architecture of Dynamic Neural Turing Machine (D-NTM). In Section 3, we describe the proposed addressing mechanism for D-NTM. Section 4 explains the training procedure. In Section 5, we briefly  discuss some related models. In Section 6, we report results on episodic question answering task. In Section 7, 8, and 9 we discuss the results in sequential MNIST, SNLI, and algorithmic toy tasks respectively. Section 10 concludes the article. 

\section{Dynamic Neural Turing Machine}

The proposed dynamic neural Turing machine (D-NTM) extends the neural Turing
machine (NTM, \citep{graves2014neural}) which has a modular design. The D-NTM
consists of two main modules: a controller, and a memory. The controller, which is
often implemented as a recurrent neural network, issues a command to the memory
so as to read, write to and erase a subset of memory cells. 

\subsection{Memory}

D-NTM consists of an external memory $\mM_t$, where each memory cell $i$ in $\mM_t[i]$ is partitioned into two parts: a trainable address vector $\mA_t[i] \in \RR^{1 \times d_a}$ and a content vector $\mC_t[i] \in \RR^{1 \times d_c}$.
\[
    \mM_t[i] = \left[ \mA_t[i] ; \mC_t[i] \right].
\]
Memory $\mM_t$ consists of $N$ such memory cells and hence represented by a rectangular matrix $\mM_t \in \RR^{N \times (d_c + d_a)}$:
\[
    \mM_t = \left[ \mA_t ; \mC_t \right].
\]
The first part $\mA_t \in \RR^{N \times d_a}$ is a learnable address matrix, and the second $\mC_t \in \RR^{N \times d_c}$ a content matrix. The address part $\mA_t$ is considered a model parameter that is updated during
training. During inference, the address part is not overwritten by the
controller and remains constant. On the other hand, the content part $\mC_t$ is
both read and written by the controller both during training and inference. At
the beginning of each episode, the content part of the memory is refreshed to be an
all-zero matrix, $\mC_0= \mathbf{0}$. This introduction of the learnable address portion for each memory cell allows the model to learn sophisticated location-based addressing strategies. 

\subsection{Controller}

At each timestep $t$, the controller (1) receives an input value $\vx_t$, (2)
addresses and reads the memory and creates the content vector $\vr_t$, (3) erases/writes a
portion of the memory, (4) updates its own hidden state $\vh_t$, and (5) outputs
a value $\vy_t$ (if needed.) In this paper, we use both a gated recurrent unit (GRU,
\citep{cho2014learning}) and a feedforward-controller to implement the controller such that for a GRU controller
\begin{align}
    \vh_t &= \text{GRU}(\vx_t, \vh_{t-1}, \vr_t) 
\end{align}
and for a feedforward-controller
\begin{align}
    \vh_t &= \sigmoid(\vx_t, \vr_t).
\end{align}

\subsection{Model Operation}

At each timestep $t$, the controller receives an input value $\vx_t$. Then it generates the read weights $\vw^r_t \in \RR^{N \times 1}$. By using the read weights $\vw^r_t$, the content vector read from the memory $\vr_t \in \RR^{(d_a + d_c) \times 1}$ is computed as
\begin{align}
    \label{eq:read_with_attention}
    \vr_t =  (\mM_{t})^{\top} \vw^r_t,
\end{align}
The hidden state of the controller ($\vh_t$) is conditioned on the memory content vector $\vr_t$ and based on this current hidden state of the controller. The model predicts the output label $\vy_t$ for the input. 

The controller also updates the memory by erasing the old content and writing a new content into the memory. The controller computes three vectors: erase vector $\ve_t \in \RR^{d_c \times 1}$, write weights $\vw^w_t \in \RR^{N \times 1}$, and candidate memory content vector $\bar{\vc}_t \in \RR^{d_c \times 1}$. These vectors are used to modify the memory. Erase vector is computed by a simple MLP which is conditioned on the hidden state of the controller $\vh_t$. The candidate memory content vector $\bar{\vc}_t$ is computed based on the current hidden state of the controller $\vh_{t} \in \RR^{d_h \times 1}$ and the input of the controller which is scaled by a scalar gate $\alpha_t$. The $\alpha_t$ is a function of the hidden state and the input of the controller. 
\begin{align}
       \alpha_t &= f(\vh_t, \vx_t), \\
        \label{eqn:cand_memory_content}
       \bar{\vc}_t &= \text{ReLU}(\mW_m \vh_t + \alpha_t \mW_x \vx_t).
\end{align}
where $\mW_m$ and $\mW_x$ are trainable matrices and $\text{ReLU}$ is the rectified linear activation function \citep{nair2010rectified}. Given the erase, write and candidate memory content vectors ($\ve_t$, $w_t^w$, and $\bar{\vc}_t$ respectively), the memory matrix is updated by,
\begin{align}
\label{eq:write_with_attention}
\mC_t[j] = (1 - \ve_t w^w_t[j]) \odot \mC_{t-1}[j] + w^w_t[j] \bar{\vc}_{t}.
\end{align}
where the index $j$ in $\mC_t[j]$ denotes the $j$-th row of the content matrix $\mC_t$ of the memory matrix $\mM_t$.

\paragraph{No Operation (NOP)}

As found in \citep{joulin2015inferring}, an additional NOP operation can be useful
for the controller {\it not} to access the memory only once in a while. We model
this situation by designating one memory cell as a NOP cell to which the controller should
access when it does not need to read or write into the memory. Because reading from or writing
into this memory cell is completely ignored. 

\begin{figure}[htbp]
\centering
\includegraphics[scale=0.50]{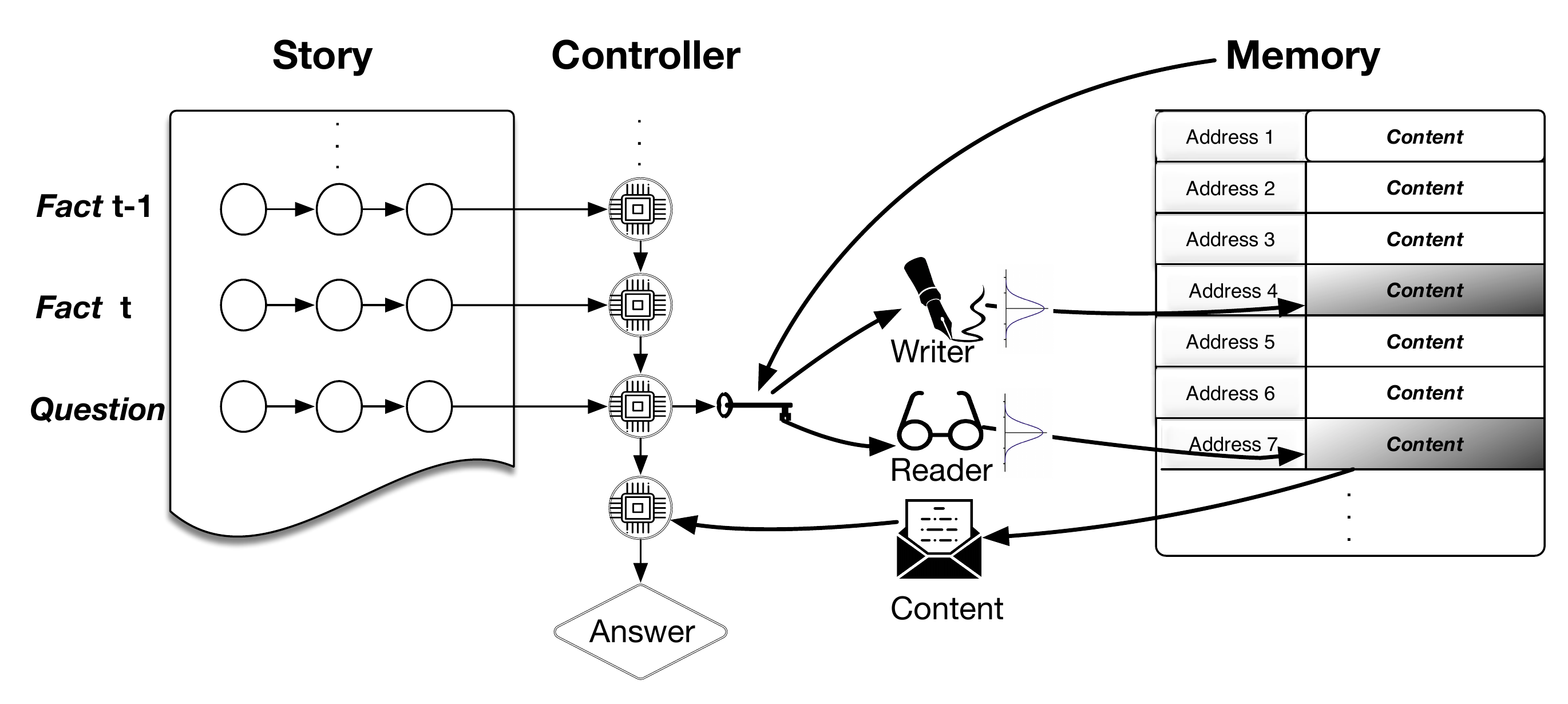}
\caption{A graphical illustration of the proposed dynamic neural Turing machine with the 
    recurrent-controller.  The controller receives the fact as a
    continuous vector encoded by a recurrent neural network, computes the
    read and write weights for addressing the memory. If the D-NTM automatically
    detects that a query has been received, it returns an answer
    and terminates.}
\label{fig_main}
\end{figure}

We illustrate and elaborate more on the read and write operations of the D-NTM in Figure \ref{fig_main}.

The computation of the read $\vw^r_t$ and write vector $\vw^w_t$ are the most crucial parts of the model since the controller decide where to read from and write into the memory by using those. We elaborate this in the next section. 

\section{Addressing Mechanism}

Each of the address vectors (both read and write) is computed in similar
ways. First, the controller computes a key vector:
\[
    \vk_t = \mW^\top_k \vh_t + \vb_k,
\]
Both for the read and the write operations, $\vk_t \in \RR^{(d_a + d_c) \times 1}$. $\mW_k \in \RR^{(d_a+d_c) \times N}$ and $\vb_k \in \RR^{(d_a + d_c) \times 1}$ are the learnable weight matrix and bias respectively of $\vk_t$.  Also, the sharpening factor $\beta_t \in \RR \ge 1$ is computed as follows:

\begin{align}
    \beta_t &= \text{softplus}(\vu_{\beta}^{\top} \vh^t + b_{\beta}) + 1.
\end{align}
where $\vu_{\beta}$ and $b_{\beta}$ are the parameters of the sharpening factor $\beta_t$ and softplus is defined as follows:
\begin{align}
\text{softplus}(x) &= \text{log}(\text{exp}(x) + 1)
\end{align}
Given the key $\vk_t$ and sharpening factor $\beta_t$, the logits for the address weights are then computed by,
\begin{align}
 z_t[i] & = \beta^t S\left(\vk_t, \mM_t[i]\right)
\end{align}
where the similarity function is basically the cosine distance where it is defined as $S\left(\vx, \vy\right) \in \RR$ and $1 \ge S\left(\vx, \vy\right) \ge -1$,
\[
    S\left(\vx, \vy\right) = \frac{\vx \cdot \vy}{||\vx||||\vy|| + \epsilon}.
\]
$\epsilon$ is a small positive value to avoid division by zero. We have used $\epsilon=1e-7$ in all our experiments. The address weight generation which we have described in this section is same with the content based addressing mechanism proposed in \citep{graves2014neural}.

\subsection{Dynamic Least Recently Used Addressing}

We introduce a memory addressing operation that can learn to put more emphasis on the 
least recently used~(LRU)~memory locations. As observed in \citep{santoro2016one, dmnew},
we find it easier to learn the write operations with the use of LRU addressing.

To learn a LRU based addressing, first we compute the exponentially moving averages of the logits~($\vz_t$) as~$\vv_t$, where it can be computed as~$\vv_t~=~0.1 \vv_{t-1} + 0.9 \vz_{t}$. We rescale the accumulated $\vv_t$ with $\gamma_t$, such that the controller adjusts the influence of how much previously written memory locations should effect the attention weights of a particular time-step. Next, we subtract $\vv_t$ from $\vz_t$ in order to reduce the weights of previously read or written memory locations. $\gamma_t$ is a shallow MLP with a scalar output and it is conditioned on the hidden state of the controller. $\gamma_t$ is parametrized with the parameters $\vu_{\gamma}$ and $\vb_{\gamma}$,
\begin{align}
       \gamma_t &= \text{sigmoid}(\vu_{\gamma}^{\top} \vh_t + \vb_{\gamma}),\\
       \vw_t &= \text{softmax}(\vz_t - \gamma_t \vv_{t-1}).  \label{eq:softmax}
\end{align}

This addressing method increases the weights of the least recently used rows of the memory. The magnitude of the influence of the least-recently used memory locations is being learned and adjusted with $\gamma_t$. Our LRU addressing is dynamic due to the model's ability to switch between pure content-based addressing and LRU. During the training, we do not backpropagate through $\vv_t$. Due to the dynamic nature of this addressing mechanism, it can be used for both read and write operations. If needed, the model will automatically learn to disable LRU while reading from the memory.

The address vector defined in Equation ~\eqref{eq:softmax} is a continuous vector. This makes the addressing operation differentiable and we refer to such a D-NTM as continuous D-NTM.

\subsection{Discrete Addressing}
\label{sec:gen_disc_add_vecs}

By definition in Eq.~\eqref{eq:softmax}, every element in the address
vector $\vw_t$ is positive and sums up to one. In other words, we can treat this vector as
the probabilities of a categorical distribution $\mathcal{C}(\vw_t)$ with $\dim(\vw_t)$ choices:
\[
    p[j] = \vw_t[j],
\]
where $\vw_t[j]$ is the $j$-th element of $\vw_t$. We can readily sample from this
categorical distribution and form an one-hot vector $\tilde{\vw}_t$ such that
\[
    \tilde{\vw}_t[k] = I(k=j),
\]
where $j \sim \mathcal{C}(\vw)$, and $I$ is an indicator function. If we use $\tilde{\vw}_t$ instead of $\vw_t$, then we will read and write from only one memory cell at a time. This makes the addressing operation non-differentiable and we refer to such a D-NTM as discrete D-NTM. In discrete D-NTM we sample the one-hot vector during training. Once training is over, we switch to a deterministic strategy. We simply choose
an element of $\vw_t$ with the largest value to be the index of the target memory
cell, such that
\[
    \tilde{\vw}_t[k] = \mathbf{I}(k=\text{argmax}(\vw_t)).
\]

\subsection{Multi-step Addressing}

At each time-step, controller may require more than one-step for accessing to the memory. 
The original NTM addresses this by implementing multiple sets of read, erase and write heads.
In this paper, we explore an option of allowing each head to operate more than once at each
timestep, similar to the multi-hop mechanism from the end-to-end memory network~\citep{sukhbaatarend}.

\section{Training D-NTM}
Once the proposed D-NTM is executed, it returns the output distribution $p(\vy^{(n)}|\vx_1^{(n)}, \ldots, \vx_T^{(n)};~\TT)$ for the $n^{th}$ example that is parameterized with $\TT$. We define our cost function as the negative log-likelihood:
\begin{align}
    \label{eq:cost}
C(\theta) = -\frac{1}{N} \sum_{n=1}^N \log p(\vy^{(n)}|\vx_1^{(n)}, \ldots, \vx_T^{(n)};~\TT),
\end{align}
where $\TT$ is a set of all the parameters of the model. 

Continuous D-NTM, just like the original NTM, is fully end-to-end differentiable and hence we can compute the gradient of this cost function by using backpropagation and learn the parameters of the model with a gradient-based optimization algorithm, such as stochastic gradient descent, to train it end-to-end. However, in discrete D-NTM, we use sampling-based strategy for all the heads during training. This clearly makes the use of backpropagation infeasible to compute the gradient, as the sampling procedure is not differentiable.

\subsection{Training discrete D-NTM}

To train discrete D-NTM, we use
REINFORCE~\citep{williams92} together with the three variance reduction
techniques--global baseline, input-dependent baseline and variance
normalization-- suggested in \citep{mnih2014neural}. 

Let us define $R(\vx)=\log p (\vy|\vx_1, \ldots, \vx_T; \TT)$ as a reward. We first center and re-scale the reward by,
\[
\tilde{R}(\vx) = \frac{R(\vx) - b}{\sqrt{\sigma^2 + \epsilon}},
\]
where $b$ and $\sigma$ is running average and standard deviation of $R$. We can further center it for each input $\vx$ separately, i.e., 
\[
\bar{R}(\vx) = \tilde{R}(\vx) - b(\vx),
\]
where $b(\vx)$ is computed by a baseline network which takes as input $\vx$ and predicts its estimated reward. The baseline network is trained to minimize the Huber loss~\citep{huber1964} between the true reward $\tilde{R}(\vx)$ and the predicted reward $b(\vx)$. This is also called as input based baseline (IBB) which is introduced in \citep{mnih2014neural}.

We use the Huber loss to learn the baseline $b(\vx)$ which is defined by,
\begin{align*}
    H_{\delta}(z) = 
    \begin{cases}
 {z^2}                   & \text{for } |z| \le \delta, \\
 \delta (2|z| - \delta), & \text{otherwise,}
\end{cases}
\end{align*}
due to its robustness where $z$ would be $\bar{R}(\vx)$ in this case.
As a further measure to reduce the variance, we regularize
the negative entropy of all those category distributions to facilitate a
better exploration during training~\citep{xu2015show}.

Then, the cost function for each training example is approximated as in Equation \eqref{eqn:reinforced_cost}. In this equation, we write the terms related to compute the REINFORCE gradients that includes terms for the entropy regularization on the action space, the likelihood-ratio term to compute the REINFORCE gradients both for the read and the write heads.
\begin{align*}
    C^n(\theta) =& -\log p(\vy|\vx_{1:T},
    \tilde{\vw}^r_{1:J}, \tilde{\vw}^w_{1:J}) \\
    & - \sum_{j=1}^J \bar{R}(\vx^n) (\log p(\tilde{\vw}^r_j|\vx_{1:T}) + \log p(\tilde{\vw}^w_j|\vx_{1:T})  \\
    & - \lambda_H \sum_{j=1}^J (\mathcal{H}(\vw^r_j|\vx_{1:T}) + \mathcal{H}(\vw^w_j|\vx_{1:T})).\numberthis \label{eqn:reinforced_cost}
\end{align*}

where $J$ is the number of addressing steps, $\lambda_H$ is the entropy regularization coefficient, and $\mathcal{H}$ denotes the entropy.

\subsection{Curriculum Learning for the Discrete Attention}
\label{sec:curr_disc_att}
Training discrete attention with feedforward controller and REINFORCE is challenging. We propose to use a curriculum strategy for training with the discrete attention in order to tackle this problem. For each minibatch, the controller stochastically decides to choose either to use the discrete or continuous weights based on the random variable $\pi_n$ with probability $p_n$ where $n$ stands for the number of $k$ minibatch updates such that we only update $p_n$ every $k$ minibatch updates. $\pi_n$ is a Bernoulli random variable which is sampled with probability of $p_n$, $\pi_n \sim \text{Bernoulli}(p_n)$. The model will either use the discrete or the continuous-attention based on the $\pi_n$. We start the training procedure with $p_0=1$ and during the training $p_n$ is annealed to $0$ by setting $p_n=\frac{p_0}{\sqrt{1+n}}$.

We can rewrite the weights $\vw_t$ as in Equation \eqref{eqn:curr_comb}, where it is expressed as the combination of continuous attention weights $\bar{\vw}_t$ and discrete attention weights $\tilde{\vw}_t$ with $\pi_t$ being a binary variable that chooses to use one of them,
\begin{equation}
    \label{eqn:curr_comb}
    \vw_t = \pi_n \bar{\vw}_t + (1-\pi_n) \tilde{\vw}_t.    
\end{equation}

By using this curriculum learning strategy, at the beginning of the training, the model learns to use the memory mainly with the continuous attention. As we anneal the $p^t$, the model will rely more on the discrete attention.

\subsection{Regularizing D-NTM}
\label{sec:reg_dntm}

If the controller of D-NTM is a recurrent neural network, we find it to be
important to regularize the training of the D-NTM so as to avoid suboptimal
solutions in which the D-NTM ignores the memory and works as a simple recurrent
neural network. 

\paragraph{Read-Write Consistency Regularizer}

One such suboptimal solution we have observed in our preliminary experiments
with the proposed D-NTM is that the D-NTM uses the address part $\mA$ of the
memory matrix simply as an additional weight matrix, rather than as a means to
accessing the content part $\mC$. We found that this pathological case can be
effectively avoided by encouraging the read head to point to a memory cell which
has also been pointed by the write head. This can be implemented as the
following regularization term:
\begin{align}
    R_{\text{rw}}(\vw^r, \vw^w) = \lambda \sum_{t'=1}^{T}||1 -
    (\frac{1}{t'}\sum_{t=1}^{t'}\vw^w_t)^{\top}\vw^r_{t'} ||_2^2
\end{align}

In the equations above, $\vw^w_t$ is the write and $\vw^r_{t}$ is the read weights.

\paragraph{Next Input Prediction as Regularization}

Temporal structure is a strong signal that should be exploited by the controller based on a recurrent neural network. We exploit this structure by letting the controller {\it predict} the input in the future. We maximize the predictability of the next input by the controller during training. This is equivalent to minimizing the following regularizer:
\[
R_{\text{pred}}(\mW) = -\sum_{t=0}^T\log p(\vx_{t+1}|\vx_t, \vw^r_t, \vw^w_t, \ve_t, \mM_t; \TT)
\]
where $\vx_t$ is the current input and $\vx_{t+1}$ is the input at the next timestep. We find this regularizer to be effective in our preliminary experiments and use it for bAbI tasks.

\section{Related Work}

A recurrent neural network (RNN), which is used as a controller in the proposed
D-NTM, has an implicit memory in the form of recurring hidden states. Even with
this implicit memory, a vanilla RNN is however known to have difficulties in
storing information for long time-spans~\citep{bengio1994learning,hochreiter1991untersuchungen}. Long short-term memory
(LSTM, \citep{lstm1997}) and gated recurrent units (GRU, \citep{cho2014learning})
have been found to address this issue. However all these models based solely on
RNNs have been found to be limited when they are used to solve, e.g.,
algorithmic tasks and episodic question-answering. 

In addition to the finite random access memory of the neural Turing machine,
based on which the D-NTM is designed, other data structures have been proposed
as external memory for neural networks. In \citep{sun1997neural,grefenstette2015learning,joulin2015inferring}, a continuous,
differentiable stack was proposed. In \citep{zaremba2015learning,rlntm}, grid and
tape storage are used. These approaches differ from the NTM in that their
memory is unbounded and can grow indefinitely. On the other hand, they are
often not randomly accessible. \cite{zhang2015structured} proposed a variation of NTM
that has a structured memory and they have shown experiments on copy and associative
recall tasks with this model.

In parallel to our work \citep{yang2016lie} and \citep{graves2016hybrid} proposed
new memory access mechanisms to improve NTM type of models. \citep{graves2016hybrid}
reported superior results on a diverse set of algorithmic learning tasks.

Memory networks~\citep{weston2014memory} form another family of neural networks
with external memory. In this class of neural networks, information is stored
explicitly as it is (in the form of its continuous representation) in the
memory, without being erased or modified during an episode. Memory networks and
their variants have been applied to various tasks
successfully~\citep{sukhbaatarend,bordes2015large,dodge2015,dmn2, chandar2016hierarchical}. \cite{keyval} 
have also independently proposed the idea of having separate key and value vectors for memory networks. A similar addressing mechanism is also explored in \citep{reed2015neural} in the context of learning program traces.

Another related family of models is the attention-based neural networks. Neural
networks with continuous or discrete attention over an input have shown promising results
on a variety of challenging tasks, including machine
translation~\citep{bahdanau2014neural,luong2015effective}, speech
recognition~\citep{chorowski2015attention}, machine reading
comprehension~\citep{hermann2015teaching} and image caption
generation~\citep{xu2015show}.  

The latter two, the memory network and attention-based networks, are however
clearly distinguishable from the D-NTM by the fact that they do not modify the
content of the memory.  



\section{Experiments on Episodic Question-Answering}

In this section, we evaluate the proposed D-NTM on the synthetic episodic
question-answering task called Facebook bAbI~\citep{weston2015towards}. We use
the version of the dataset that contains 10k training examples per sub-task provided by
Facebook.\footnote{
    \url{https://research.facebook.com/researchers/1543934539189348}
} For each episode, the D-NTM reads a sequence of factual sentences followed by a
question, all of which are given as natural language sentences. The D-NTM is
expected to store and retrieve relevant information in the memory in order to
answer the question based on the presented facts. 

\subsection{Model and Training Details}

We use the same hyperparameters for all the tasks for a given model. We use a recurrent neural network with GRU units to encode a variable-length
fact into a fixed-size vector representation. This allows the D-NTM to exploit
the word ordering in each fact, unlike when facts are encoded as bag-of-words
vectors. We experiment with both a recurrent and feedforward neural network as the controller that generates the read and write weights. The controller has 180 units.
We train our feedforward controller using noisy-tanh activation function~\citep{gulcehre2016noisy} since we were experiencing training difficulties with $\text{sigmoid}$ and $\text{tanh}$ activation functions. We use both single-step and three-steps addressing with our GRU controller. The memory contains 120 memory cells. Each memory cell consists of a 16-dimensional address part and 28-dimensional content part.

We set aside a random $10\%$ of the training examples as a validation set for each
sub-task and use it for early-stopping and hyperparameter search. We train one
D-NTM for each sub-task, using Adam~\citep{adam} with its learning rate set to
$0.003$ and $0.007$ respectively for GRU and feedforward controller. The size of each 
minibatch is 160, and each minibatch is constructed uniform-randomly from the training set.

\subsection{Goals}

The goal of this experiment is three-fold. First, we present for the first time the performance of a memory-based network that can {\it both} read and write dynamically on the Facebook bAbI tasks\footnote{Similar experiments were done in the recently published \citep{graves2016hybrid}, but D-NTM results for bAbI tasks were already available in arxiv by that time.}. We aim to understand whether a model that has to learn to write an incoming fact to the memory, rather than storing it as it is, is able to work well, and to do so, we compare both the original NTM and proposed D-NTM against an LSTM-RNN. 

Second, we investigate the effect of having to learn how to write. The fact that the NTM needs to learn to write likely has adverse effect on the overall performance, when compared to, for instance, end-to-end memory networks (MemN2N, \citep{sukhbaatarend}) and dynamic memory network (DMN+, \citep{dmn2}) both of which simply store the incoming facts as they are. We quantify this effect in this experiment. Lastly, we show the effect of the proposed learnable addressing scheme. 

We further explore the effect of using a feedforward controller instead of the GRU controller. In addition to the explicit memory, the GRU controller can use its own internal hidden state as the memory. On the other hand, the feedforward controller must solely rely on the explicit memory, as it is the only memory available.

\begin{table*}[htbp]
\vspace{-2mm}
  \centering
  \tiny 
\begin{tabular}{ | l || c | c | c || c |c |c|c|| c | c | c|c| }
\hline
& & & & 1-step & 1-step & 1-step & 1-step & 3-steps & 3-steps & 3-steps & 3-steps \\ 
& & & & LBA$^{\ast}$ & CBA & Soft & Discrete & LBA$^{\ast}$ & CBA & Soft & Discrete\\
Task & LSTM & MemN2N & DMN+ & NTM & NTM & D-NTM & D-NTM  & NTM & NTM & D-NTM & D-NTM \\ \hline

1   &   0.00    &   0.00    &   0.00    &   16.30   &   16.88   &   5.41    &   6.66    &   0.00    &   0.00    &   0.00    &   0.00\\   
2   &   81.90   &   0.30    &   0.30    &   57.08   &   55.70   &   58.54   &   56.04   &   61.67   &   59.38   &   46.66   &   62.29 \\ 
3   &   83.10   &   2.10    &   1.10    &   74.16   &   55.00   &   74.58   &   72.08   &   83.54   &   65.21   &   47.08   &   41.45  \\
4   &   0.20    &   0.00    &   0.00    &   0.00    &   0.00    &   0.00    &   0.00    &   0.00    &   0.00    &   0.00    &   0.00   \\
5   &   1.20    &   0.80    &   0.50    &   1.46    &   20.41   &   1.66    &   1.04    &   0.83    &   1.46    &   1.25    &   1.45   \\
6   &   51.80   &   0.10    &   0.00    &   23.33   &   21.04   &   40.20   &   44.79   &   48.13   &   54.80   &   20.62   &   11.04  \\
7   &   24.90   &   2.00    &   2.40    &   21.67   &   21.67   &   19.16   &   19.58   &   7.92    &   37.70   &   7.29    &   5.62   \\
8   &   34.10   &   0.90    &   0.00    &   25.76   &   21.05   &   12.58   &   18.46   &   25.38   &   8.82    &   11.02   &   0.74   \\
9   &   20.20   &   0.30    &   0.00    &   24.79   &   24.17   &   36.66   &   34.37   &   37.80   &   0.00    &   39.37   &   32.50  \\
10  &   30.10   &   0.00    &   0.00    &   41.46   &   33.13   &   52.29   &   50.83   &   56.25   &   23.75   &   20.00   &   20.83  \\
11  &   10.30   &   0.10    &   0.00    &   18.96   &   31.88   &   31.45   &   4.16    &   3.96    &   0.28    &   30.62   &   16.87  \\
12  &   23.40   &   0.00    &   0.00    &   25.83   &   30.00   &   7.70    &   6.66    &   28.75   &   23.75   &   5.41    &   4.58   \\
13  &   6.10    &   0.00    &   0.00    &   6.67    &   5.63    &   5.62    &   2.29    &   5.83    &   83.13   &   7.91    &   5.00   \\
14  &   81.00   &   0.10    &   0.20    &   58.54   &   59.17   &   60.00   &   63.75   &   61.88   &   57.71   &   58.12   &   60.20  \\
15  &   78.70   &   0.00    &   0.00    &   36.46   &   42.30   &   36.87   &   39.27   &   35.62   &   21.88   &   36.04   &   40.26  \\
16  &   51.90   &   51.80   &   45.30   &   71.15   &   71.15   &   49.16   &   51.35   &   46.15   &   50.00   &   46.04   &   45.41  \\
17  &   50.10   &   18.60   &   4.20    &   43.75   &   43.75   &   17.91   &   16.04   &   43.75   &   56.25   &   21.25   &   9.16   \\
18  &   6.80    &   5.30    &   2.10    &   3.96    &   47.50   &   3.95    &   3.54    &   47.50   &   47.50   &   6.87    &   1.66   \\
19  &   90.30   &   2.30    &   0.00    &   75.89   &   71.51   &   73.74   &   64.63   &   61.56   &   63.65   &   75.88   &   76.66  \\
20  &   2.10    &   0.00    &   0.00    &   1.25    &   0.00    &   2.70    &   3.12    &   0.40    &   0.00    &   3.33    &   0.00   \\\hline

Avg.Err. & 36.41 & 4.24 & \textbf{2.81} & 31.42 & 33.60 & 29.51 & \textbf{27.93} & 32.85 & 32.76 & 24.24 & \textbf{21.79} \\\hline
\end{tabular}
\caption{Test error rates (\%) on the 20 bAbI QA tasks for models using 10k training examples with the GRU and feedforward controller. FF stands for the experiments that are conducted with feedforward controller. Let us, note that LBA$^{\ast}$ refers to NTM that uses both LBA and CBA. In this table, we compare multi-step vs single-step addressing, original NTM with location based+content based addressing vs only content based addressing, and discrete vs continuous addressing on bAbI.}
\label{app:babi_10k_soft}
\end{table*}

\subsection{Results and Analysis}

In Table~\ref{app:babi_10k_soft}, we first observe that the NTMs are indeed capable of solving this type of episodic question-answering better than the vanilla LSTM-RNN. Although the availability of explicit memory in the NTM has already suggested this result, we note that this is the first time neural Turing machines have been used in this specific task.

All the variants of NTM with the GRU controller outperform the vanilla LSTM-RNN. However, not all of them perform equally well. First, it is clear that the proposed dynamic NTM (D-NTM) using the GRU controller outperforms the original NTM with the GRU controller (NTM, CBA only NTM vs. continuous D-NTM, Discrete D-NTM). As discussed earlier, the learnable addressing scheme of the D-NTM allows the controller to access the memory slots by location in a potentially nonlinear way. We expect it to help with tasks that have non-trivial access patterns, and as anticipated, we see a large gain with the D-NTM over the original NTM in the tasks of, for instance, 12 - Conjunction and 17 - Positional Reasoning. 

Among the recurrent variants of the proposed D-NTM, we notice significant improvements by using discrete addressing over using continuous addressing. We conjecture that this is due to certain types of tasks that require precise/sharp retrieval of a stored fact, in which case continuous addressing is in disadvantage over discrete addressing. This is evident from the observation that the D-NTM with discrete addressing significantly outperforms that with continuous addressing in the tasks of 8 - Lists/Sets and 11 - Basic Coreference. Furthermore, this is in line with an earlier observation in \citep{xu2015show}, where discrete addressing was found to generalize better in the task of image caption generation. 

In Table~\ref{app:babi_10k_soft2}, we also observe that the D-NTM with the feedforward controller and discrete attention performs worse than LSTM and D-NTM with continuous-attention. However, when the proposed curriculum strategy from Sec.~\ref{sec:gen_disc_add_vecs} is used, the average test error drops from 68.30 to 37.79.

We empirically found training of the feedforward controller more difficult than that of the recurrent controller. We train our feedforward controller based models four times longer (in terms of the number of updates) than the recurrent controller based ones in order to ensure that they are converged for most of the tasks. On the other hand, the models trained with the GRU controller overfit on bAbI tasks very quickly. For example, on tasks 3 and 16 the feedforward controller based model underfits~(i.e., high training loss) at the end of the training, whereas with the same number of units the model with the GRU controller can overfit on those tasks after 3,000 updates only.

We notice a significant performance gap, when our results are compared to the variants of the memory network~\citep{weston2014memory} (MemN2N and DMN+). We attribute this gap to the difficulty in learning to manipulate and store a complex input. 

\cite{graves2016hybrid} also has also reported results with differentiable neural computer (DNC) and NTM on bAbI dataset. However their experimental setup is different from the setup we use in this paper. This makes the comparisons between more difficult. The main differences broadly are, as the input representations to the controller, they used the embedding representation of each word whereas we have used the representation obtained with GRU for each fact. Secondly, they report only joint training results. However, we have only trained our models on the individual tasks separately. However, despite the differences in terms of architecture in DNC paper (see Table 1), the mean results of their NTM results is very close to ours 28.5\% with std of +/- 2.9 which we obtain 31.4\% error.

\begin{table*}[htbp]
\vspace{-2mm}
  \centering
  \tiny 
\begin{tabular}{ |c|c|c|c| }
\hline
& FF & FF & FF\\ 
& Soft & Discrete & Discrete$^*$\\
Task &  D-NTM & D-NTM & D-NTM\\ \hline
1   &   4.38    &   81.67   &   14.79\\
2   &   27.5    &   76.67   &   76.67\\
3   &   71.25   &   79.38   &   70.83\\
4   &   0.00    &   78.65   &   44.06\\
5   &   1.67    &   83.13   &   17.71\\
6   &   1.46    &   48.76   &   48.13\\
7   &   6.04    &   54.79   &   23.54\\
8   &   1.70    &   69.75   &   35.62\\
9   &   0.63    &   39.17   &   14.38\\
10  &   19.80   &   56.25   &   56.25\\
11  &   0.00    &   78.96   &   39.58\\
12  &   6.25    &   82.5    &   32.08\\
13  &   7.5     &   75.0    &   18.54\\
14  &   17.5    &   78.75   &   24.79\\
15  &   0.0     &   71.42   &   39.73\\
16  &   49.65   &   71.46   &   71.15\\
17  &   1.25    &   43.75   &   43.75\\
18  &   0.24    &   48.13   &   2.92\\
19  &   39.47   &   71.46   &   71.56\\
20  &   0.0     &   76.56   &   9.79\\\hline
Avg.Err. & \textbf{12.81} & 68.30 & 37.79\\\hline
\end{tabular}
\caption{Test error rates (\%) on the 20 bAbI QA tasks for models using 10k training examples with feedforward controller.}
\label{app:babi_10k_soft2}
\end{table*}

\subsection{Visualization of Discrete Attention}

\begin{figure}[htbp]
\centering
\includegraphics[width=0.85\textwidth]{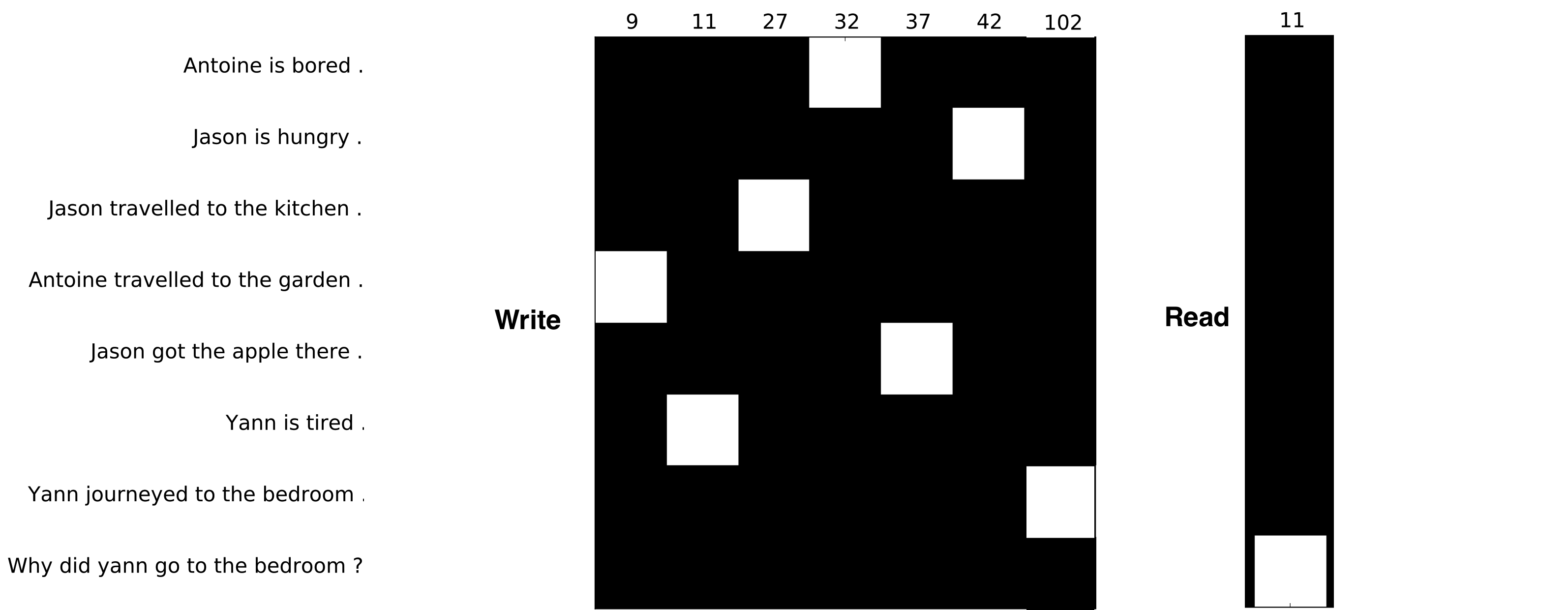}
\caption{An example view of the discrete attention over the memory slots for both read (left)  and write heads(right). x-axis the denotes the memory locations that are being accessed and y-axis corresponds to the content in the particular memory location. In this figure, we visualize the discrete-attention model with 3 reading steps and on task 20. It is easy to see that the NTM with discrete-attention accesses to the relevant part of the memory. We only visualize the last-step of the three steps for writing. Because with discrete attention usually the model just reads the empty slots of the memory.}
\label{fig:ntm_hard_att}

\end{figure}

We visualize the attention of D-NTM with GRU controller with discrete attention in Figure \ref{fig:ntm_hard_att}. From this example, we can see that D-NTM has learned to find the correct supporting fact even without any supervision for the particular story in the visualization. 

\subsection{Learning Curves for the Recurrent Controller}
\begin{figure}[htbp]
\centering
\includegraphics[width=0.8\linewidth]{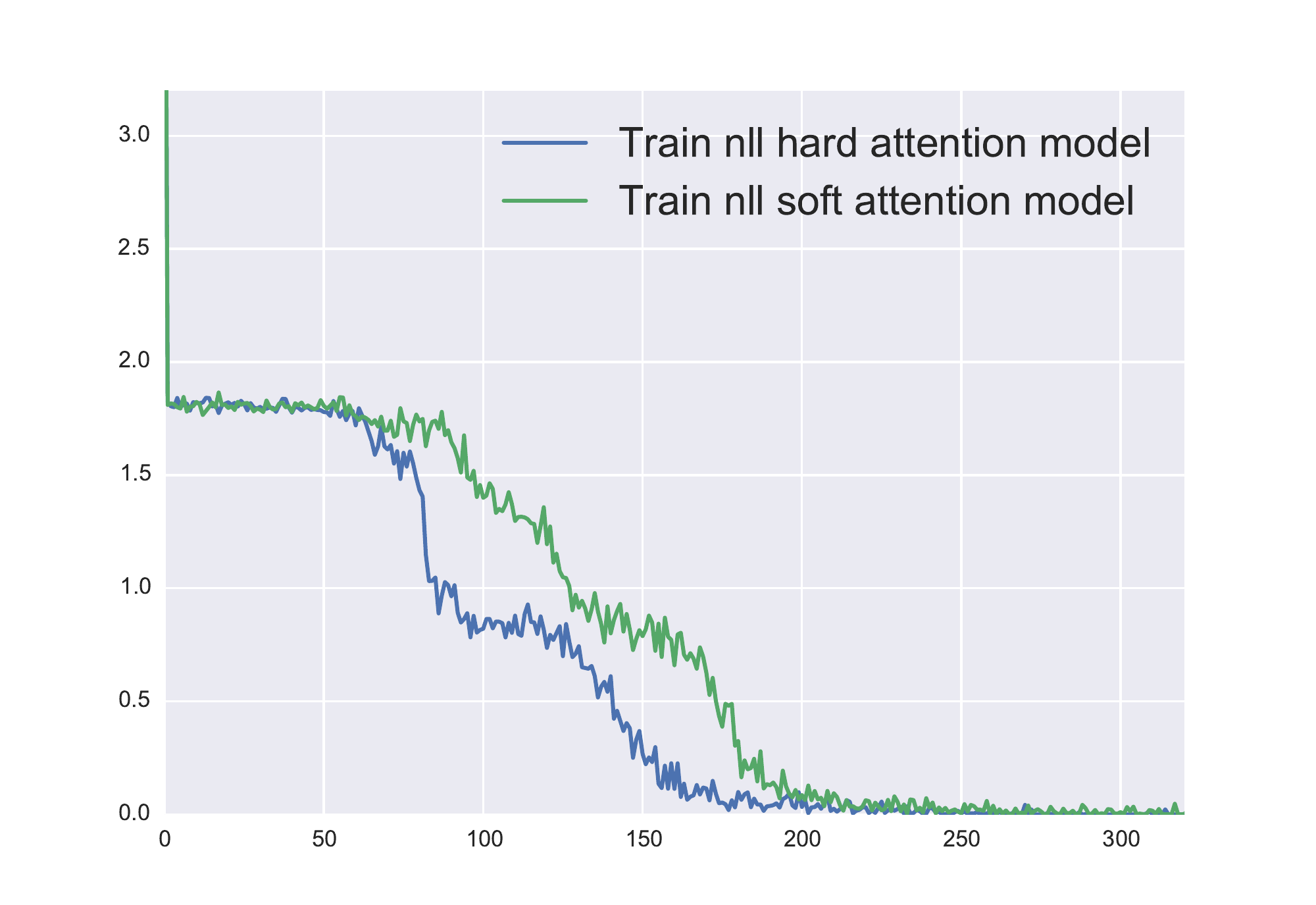}
\caption{A visualization for the learning curves of continuous and discrete D-NTM models trained on Task 1 using 3 steps. In most tasks, we observe that the discrete attention model with GRU controller does converge faster than the continuous-attention model.}
\label{fig:ntm_learn_curves}
\end{figure}

In Figure \ref{fig:ntm_learn_curves}, we compare the learning curves of the continuous and discrete attention D-NTM model with recurrent controller on Task 1. Surprisingly, the discrete attention D-NTM converges faster than the continuous-attention model. The main difficulty of learning continuous-attention is due to the fact that learning to write with continuous-attention can be challenging.

\subsection{Training with Continuous Attention and Testing with Discrete Attention}

In Table \ref{app:babi_10k_soft_ff_hard_test}, we provide results to investigate the effects of using discrete attention model at the test-time for a model trained with feedforward controller and continuous attention. Discrete$^{\ast}$ D-NTM model bootstraps the discrete attention with the continuous attention, using the curriculum method that we have introduced in Section \ref{sec:curr_disc_att}. Discrete$^{\dagger}$ D-NTM model is the continuous-attention model which uses discrete-attention at the test time. We observe that the Discrete$^{\dagger}$ D-NTM model which is trained with continuous-attention outperforms Discrete D-NTM model.

\begin{table*}[htbp]
\vspace{-2mm}
  \centering
  \footnotesize 
\begin{tabular}{ | l || c |c | c | c | c |}
\hline
&  continuous & Discrete & Discrete$^\ast$ & Discrete$^\dagger$  \\
Task & D-NTM & D-NTM  & D-NTM & D-NTM\\ \hline

1 &4.38 & 81.67 & 14.79 & 72.28\\
2 & 27.5 & 76.67 & 76.67 & 81.67 \\
3 & 71.25 & 79.38 & 70.83 & 78.95 \\
4 &0.00 & 78.65 & 44.06 & 79.69 \\
5 &1.67 & 83.13 & 17.71 & 68.54 \\
6 & 1.46 & 48.76 & 48.13 & 31.67 \\
7 & 6.04 & 54.79 & 23.54 & 49.17 \\
8 & 1.70 & 69.75 & 35.62 & 79.32 \\
9 & 0.63 & 39.17 & 14.38 & 37.71 \\
10 & 19.80 & 56.25 & 56.25 & 25.63 \\
11 & 0.00 & 78.96 & 39.58 & 82.08 \\
12 & 6.25 & 82.5 & 32.08 & 74.38 \\
13  & 7.5 & 75.0 & 18.54 & 47.08 \\
14 & 17.5 & 78.75 & 24.79 & 77.08 \\
15 & 0.0 & 71.42 & 39.73 & 73.96 \\
16 & 49.65 & 71.46 & 71.15 & 53.02 \\
17  & 1.25 & 43.75 & 43.75 & 30.42 \\
18 & 0.24 & 48.13 & 2.92 & 11.46\\
19 & 39.47 & 71.46 & 71.56 & 76.05\\
20 & 0.0 & 76.56 & 9.79 & 13.96 \\\hline
Avg & \textbf{12.81} & 68.30 & 37.79 & 57.21 \\\hline

\end{tabular}
\caption{Test error rates (\%) on the 20 bAbI QA tasks for models using 10k training examples with the feedforward controller. Discrete$^{\ast}$ D-NTM model bootstraps the discrete attention with the continuous attention, using the curriculum method that we have introduced in Section \ref{sec:gen_disc_add_vecs}. Discrete$^{\dagger}$ D-NTM model is the continuous-attention model which uses discrete-attention at the test time.}
\label{app:babi_10k_soft_ff_hard_test}
\end{table*}

\subsection{D-NTM with BoW Fact Representation}
In Table \ref{app:babi_10k_soft_bow}, we provide results for D-NTM using BoW with positional encoding~(PE) \cite{sukhbaatarend} as the representation of the input facts. The facts representations are provided as an input to the GRU controller. In agreement to our results with the 
GRU fact representation, with the BoW fact representation we observe improvements with multi-step of addressing over single-step and discrete addressing over continuous addressing.

\begin{table*}[htbp]
\vspace{-2mm}
  \centering
  \footnotesize 
\begin{tabular}{ | l || c |c | c | c | c |}
\hline
&  Soft & Discrete & Soft & Discrete  \\
Task & D-NTM(1-step) & D-NTM(1-step)  & D-NTM(3-steps) & D-NTM(3-steps)\\ \hline

1 & 0.00 & 0.00 & 0.00 & 0.00\\
2 & 61.04 & 59.37 & 56.87 & 55.62 \\
3 & 55.62 & 57.5 & 62.5 & 57.5 \\
4 & 27.29 & 24.89 & 26.45 & 27.08 \\
5 & 13.55 & 12.08 & 15.83 & 14.78 \\
6 & 13.54 & 14.37 & 21.87 & 13.33 \\
7 & 8.54 & 6.25 & 8.75 & 14.58 \\
8 & 1.69 & 1.36 & 3.01 & 3.02 \\
9 & 17.7 & 16.66 & 37.70 & 17.08 \\
10 & 26.04 & 27.08 & 26.87 & 23.95 \\
11 & 20.41 & 3.95 & 2.5 & 2.29 \\
12 & 0.41 & 0.83 & 0.20 & 4.16 \\
13  & 3.12 & 1.04 & 4.79 & 5.83 \\
14 & 62.08 & 58.33 & 61.25 & 60.62 \\
15 & 31.66 & 26.25 & 0.62 & 0.05 \\
16 & 54.47 & 48.54 & 48.95 & 48.95 \\
17  & 43.75 & 31.87 & 43.75 & 30.62 \\
18 & 33.75 & 39.37 & 36.66 & 36.04\\
19 & 64.63 & 69.21 & 67.23 & 65.46\\
20 & 1.25 & 0.00 & 1.45 & 0.00 \\\hline
Avg & 27.02 & 24.98 & 26.36 & {\bf 24.05} \\\hline

\end{tabular}
\caption{Test error rates (\%) on the 20 bAbI QA tasks for models using 10k training examples with the GRU controller and representations of facts are obtained with BoW using positional encoding.}
\label{app:babi_10k_soft_bow}
\end{table*}

\section{Experiments on Sequential $p$MNIST}

In sequential MNIST task, the pixels of the MNIST digits are provided to the model in scan line order, left to right and top to bottom \citep{le2015simple}. At the end of sequence of pixels, the model predicts the label of the digit in the sequence of pixels. We experiment D-NTM on the variation of sequential MNIST where the order of the pixels is randomly shuffled, we call this task as permuted MNIST ($p$MNIST). An important contribution of this task to our paper, in particular, is to measure the model's ability to perform well when dealing with long-term dependencies. We report our results in Table \ref{tbl:pmnist_dntm_comp}, we observe improvements over other models that we compare against. In Table \ref{tbl:pmnist_dntm_comp}, "discrete addressing with MAB" refers to D-NTM model using REINFORCE with baseline computed from moving averages of the reward. Discrete addressing with IB refers to D-NTM using REINFORCE with input-based baseline.

\begin{table}[htbp]
\centering
\footnotesize
\begin{tabular}{@{}ll@{}}
\toprule
                                                    & Test  Acc \\ \midrule
D-NTM discrete MAB     & 89.6   \\
D-NTM discrete  IB      &  92.3  \\ 
Soft D-NTM                          & \textbf{93.4}           \\
NTM                                               & 90.9          \\\midrule
I-RNN~\citep{le2015simple}                                & 82.0           \\
Zoneout~\citep{krueger2016zoneout}                                             & 93.1           \\
LSTM~\citep{krueger2016zoneout}                                               & 89.8           \\
Unitary-RNN~\citep{arjovsky2015unitary}                  & 91.4           \\
Recurrent Dropout~\citep{krueger2016zoneout}                                   & 92.5           \\ 
Recurrent Batch Normalization~\citep{cooijmans2016recurrent}    & 95.6 \\ \bottomrule
\end{tabular}
\caption{ Sequential $p$MNIST.}
\label{tbl:pmnist_dntm_comp}

\end{table}

\begin{figure}[htbp]
\centering
\includegraphics[width=0.8\linewidth]{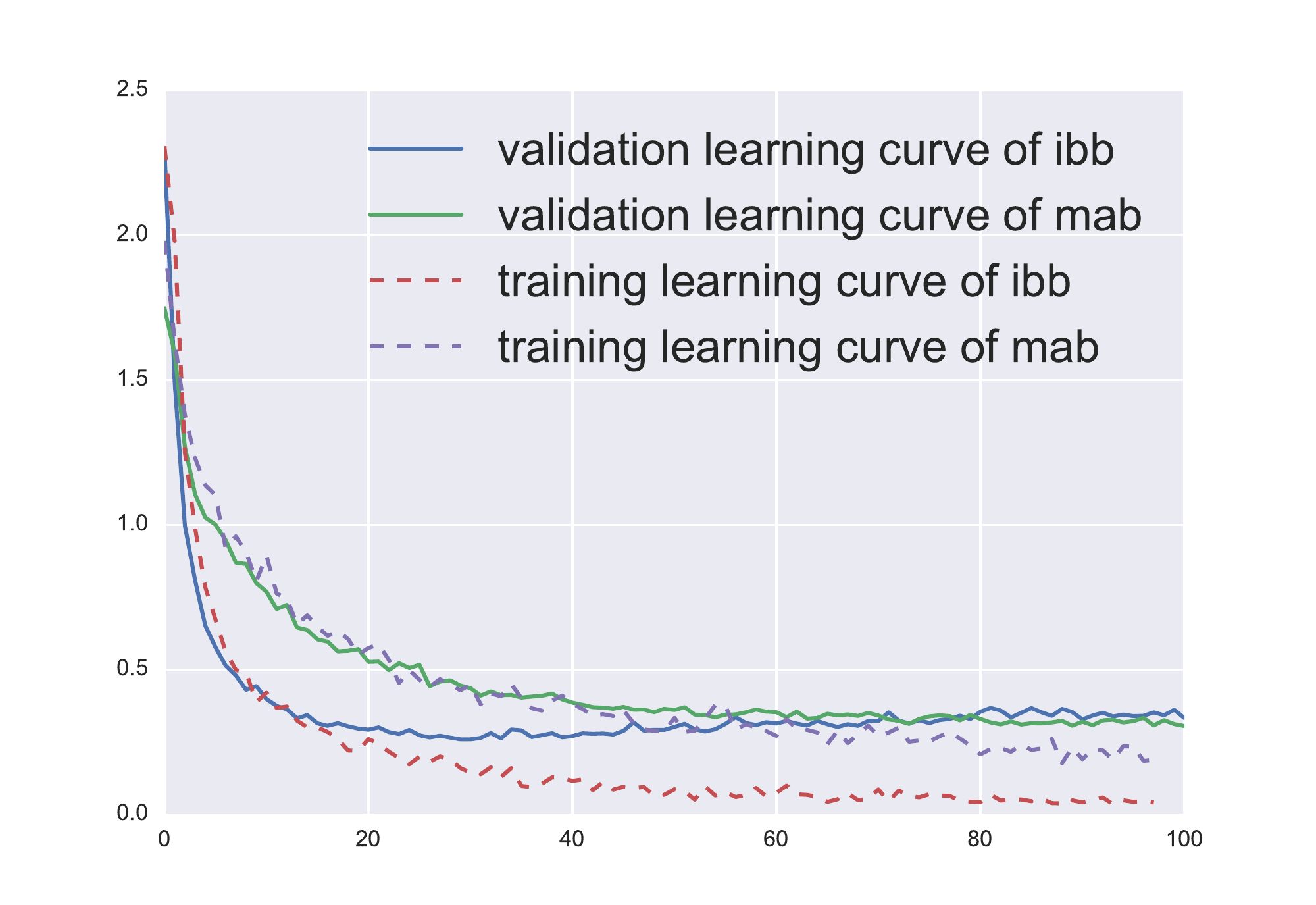}
\caption{We compare the learning curves of our D-NTM model using discrete attention on $p$MNIST task with input-based baseline and regular REINFORCE baseline. The x-axis is the loss and y-axis is the number of epochs.}
\label{fig:dntm_reinforce_baselines_cmp}
\end{figure}

In Figure \ref{fig:dntm_reinforce_baselines_cmp}, we show the learning curves of input-based-baseline (ibb) and regular REINFORCE with moving averages baseline (mab) on the $p$MNIST task. We observe that input-based-baseline in general is much easier to optimize and converges faster as well. But it can quickly overfit to the task as well. Let us note that, recurrent batch normalization with LSTM \citep{cooijmans2016recurrent} with 95.6\% accuracy and it performs much better than other algorithms. However, it is possible to use recurrent batch normalization in our model and potentially improve our results on this task as well.

In all our experiments on sequential MNIST task, we try to keep the capacity of our model to be close to our baselines. We use  100 GRU units in the controller and each content vector of size 8 and with address vectors of size 8. We use a learning rate of $1e-3$ and trained the model with Adam optimizer. We did not use the read and write consistency regularization in any of our models.

\section{Stanford Natural Language Inference~(SNLI) Task}

SNLI task \citep{bowman2015large} is designed to test the abilities of different machine learning algorithms for inferring the entailment between two different statements. Those two statements, can either entail, contradict or be neutral to each other. In this paper, we feed the premise followed by the end of premise~(EOP) token and the hypothesis in the same sequence as an input to the model. Similarly \cite{rocktaschel2015reasoning} have trained their model by providing the premise and the hypothesis in a similar way. This ensures that the performance of our model does not rely only on a particular preprocessing or architectural engineering. But rather we mainly rely on the model's ability to represent the sequence and the dependencies in the input sequence efficiently. The model proposed by \cite{rocktaschel2015reasoning}, applies attention over its previous hidden states over premise when it reads the hypothesis.

In Table \ref{tbl:snli_ntm_res}, we report results for different models with or without recurrent dropout \citep{semeniuta2016recurrent} and layer normalization \citep{ba2016layer}.

The number of input vocabulary we use in our paper is $41200$, we use GLOVE~\citep{pennington2014glove} embeddings to initialize the input embeddings. We use GRU-controller with 300 units and the size of the embeddings are also 300. We optimize our models with Adam. We have done a hyperparameter search to find the optimal learning rate via random search and sampling the learning rate from log-space between $1e-2$ and $1e-4$ for each model. We use layer-normalization in our controller \citep{ba2016layer}. 

We have observed significant improvements by using layer normalization and dropout on this task. Mainly because that the overfitting is a severe problem on SNLI. D-NTM achieves better performance compared to both LSTM and NTMs.

\begin{table}[htbp]
\centering
\footnotesize
\begin{tabular}{@{}ll@{}}
\toprule
                                                    & Test  Acc \\ \midrule
                                                                \hline
            Word by Word Attention\citep{rocktaschel2015reasoning} & \textbf{83.5} \\
            Word by Word Attention two-way\citep{rocktaschel2015reasoning} & 83.2 \\
            \hline
            LSTM + LayerNorm + Dropout & 81.7 \\
            NTM + LayerNorm + Dropout & 81.8 \\
            DNTM + LayerNorm + Dropout & \textbf{82.3} \\
            LSTM \citep{bowman2015large} & 77.6 \\
            D-NTM & 80.9 \\
            NTM & 80.2 \\ \bottomrule
\end{tabular}
\caption{ Stanford Natural Language Inference Task}
\label{tbl:snli_ntm_res}

\end{table}

\section{NTM Toy Tasks}

We explore the possibility of using D-NTM to solve algorithmic tasks such as copy and associative recall tasks. We train our model on the same lengths of sequences that is experimented in \citep{graves2014neural}. We report our results in Table \ref{tbl:ntm_toy_tasks}. We find out that D-NTM using continuous-attention can successfully learn the "Copy" and "Associative Recall" tasks.

In Table \ref{tbl:ntm_toy_tasks}, we train our model on sequences of the same length as the experiments in \citep{graves2014neural} and test the model on the sequences of the maximum length seen during the training. We consider a model to be successful on copy or associative recall if its validation cost (binary cross-entropy) is lower than $0.02$ over the sequences of maximum length seen during the training. We set the threshold to $0.02$ to determine whether a model is successful on a task. Because empirically we observe that the models have higher validation costs perform badly in terms of generalization over the longer sequences. "D-NTM discrete" model in this table is trained with REINFORCE using moving averages to estimate the baseline.

\begin{table}[ht!]

\centering
\footnotesize
\begin{tabular}{@{}lll@{}}
\toprule
           & Copy Tasks & Associative Recall \\ \midrule
Soft D-NTM & Success    & Success            \\
D-NTM discrete & Success    & Failure            \\
NTM        & Success     & Success            \\ \bottomrule
\end{tabular}
\caption{NTM Toy Tasks.}
\label{tbl:ntm_toy_tasks}
\end{table}

On both copy and associative recall tasks, we try to keep the capacity of our model to be close to our baselines. We use 100 GRU units in the controller and each content vector of has a size of 8 and using  address vector of size 8. We use a learning rate of $1e-3$ and trained the model with Adam optimizer. We did not use the read and write consistency regularization in any of our models. For the model with the discrete attention we use REINFORCE with baseline computed using moving averages.

\section{Conclusion and Future Work}

In this paper we extend neural Turing machines (NTM) by introducing a learnable addressing scheme which allows the NTM to be capable of performing highly nonlinear location-based addressing. This extension, to which we refer by dynamic NTM (D-NTM), is extensively tested with various configurations, including different addressing mechanisms (continuous vs. discrete) and different number of addressing steps, on the Facebook bAbI tasks. This is the first time an NTM-type model was tested on this task, and we observe that the NTM, especially the proposed D-NTM, performs better than vanilla LSTM-RNN. Furthermore, the experiments revealed that the discrete, discrete addressing works better than the continuous addressing with the GRU controller, and our analysis reveals that this is the case when the task requires precise retrieval of memory content. 

Our experiments show that the NTM-based models can be weaker than other variants of memory networks which do not learn but have an explicit mechanism of storing incoming facts as they are. We conjecture that this is due to the difficulty in learning how to write, manipulate and delete the content of memory. Despite this difficulty, we find the NTM-based approach, such as the proposed D-NTM, to be a better, future-proof approach, because it can scale to a much longer horizon (where it becomes impossible to explicitly store all the experiences.)

On $p$MNIST task, we show that our model can outperform other similar type of approaches proposed to deal with the long-term dependencies. On copy and associative recall tasks, we show that our model can solve the algorithmic problems that are proposed to solve with NTM type of models.

Finally we have shown some results on the SNLI task where our model performed better than NTM and the LSTM on this task. However our results do not involve any task specific modifications and the results can be improved further by structuring the architecture of our model according to the SNLI task.

The success of both the learnable address and the discrete addressing scheme suggests two future research directions. First, we should try both of these schemes in a wider array of memory-based models, as they are not specific to the neural Turing machines. Second, the proposed D-NTM needs to be evaluated on a diverse set of applications, such as text summarization~\citep{para}, visual question-answering~\citep{vqa} and machine translation, in order to make a more concrete conclusion.

\bibliographystyle{plainnat}
\bibliography{refs}

\end{document}